\documentclass[conference]{IEEEtran}

\usepackage{cite}

\usepackage{amsmath,amssymb,amsfonts}
\usepackage{algorithmic}
\usepackage{graphicx}
\usepackage{textcomp}
\usepackage{xcolor}

\usepackage[nolist]{acronym}
\usepackage{url}

\usepackage{tabularx}

\usepackage{subfigure}

\usepackage{multirow}

\def\BibTeX{{\rm B\kern-.05em{\sc i\kern-.025em b}\kern-.08em
    T\kern-.1667em\lower.7ex\hbox{E}\kern-.125emX}}
\begin{document}

%\ninept

\begin{acronym}

\acro{SDR}[SDR]{Software-Defined Radio}
\acro{DSP}[DSP]{Digital Signal Processing}
\acro{CR}[CR]{Cognitive Radio}
\acro{RF}[RF]{Radio Frequency}
\acro{AF}[AF]{Audio Frequency}

\acro{HF}[HF]{High Frequency}
\acro{VHF}[VHF]{Very High Frequency}
\acro{UHF}[UHF]{Ultra High Frequency}

\acro{PSK}[PSK]{Phase-Shift Keying}
\acro{FSK}[FSK]{Frequency-Shift Keying}
\acro{MFSK}[MFSK]{Multiple Frequency-Shift Keying}
\acro{OFDM}[OFDM]{Orthogonal Frequency-Division Multiplexing}
\acro{CW/Morse}[CW/Morse]{Continuous Wave}
\acro{CW}[CW]{Continuous Wave}
\acro{MC}[MC]{Multi-Carrier}

\acro{USB}[USB]{Upper Side Band}

\acro{BNetzA}[BNetzA]{Federal Network Agency}
\acro{FCC}[FCC]{Federal Communications Commission}

\acro{OM}[OM]{operating mode}
\acro{OMP}[OMP]{operating mode parameters}

\acro{AMC}[AMC]{Automatic Modulation Classification}

\acro{RN-18}[RN-18]{ResNet-18}
\acro{EN-B0}[EN-B0]{EfficientNetB0}
\acro{Vim-Ti}[Vim-Ti]{Vision Mamba Tiny}

\acro{SGD}[SGD]{Stochastic Gradient Descent}
\acro{ML}[ML]{Machine Learning}

\acro{SNR}[SNR]{Signal-to-Noise Ratio}

\acro{FFT}[FFT]{Fast Fourier Transform}

\end{acronym}

\title{Digital Operating Mode Classification of Real-World Amateur Radio Transmissions}

\author{
    \IEEEauthorblockN{Maximilian Bundscherer, Thomas H. Schmitt, Ilja Baumann and Tobias Bocklet}
    \IEEEauthorblockA{\textit{Technische Hochschule Nürnberg Georg Simon Ohm}\\
    Nuremberg, Germany\\
    \{firstname.lastname\}@th-nuernberg.de}
}

\maketitle

\begin{abstract}
This study presents an ML approach for classifying digital radio operating modes evaluated on real-world transmissions.
We generated  \(98\) different parameterized radio signals from \(17\) digital operating modes, transmitted each of them on the \(70\) cm (UHF) amateur radio band, and recorded our transmissions with two different architectures of SDR receivers.
Three lightweight ML models were trained exclusively on spectrograms of limited non-transmitted signals with random characters as payloads.
This training involved an online data augmentation pipeline to simulate various radio channel impairments.
Our best model, EfficientNetB0, achieved an accuracy of \(93.80\)\% across the \(17\) operating modes and \(85.47\)\% across all \(98\) parameterized radio signals, evaluated on our real-world transmissions with Wikipedia articles as payloads.
Furthermore, we analyzed the impact of varying signal durations \& the number of FFT bins on classification, assessed the effectiveness of our simulated channel impairments, and tested our models across multiple simulated SNRs.
\end{abstract}


\section{Introduction}

The ability to quickly and accurately identify primary users and other participants is essential, especially for \ac{CR} applications where several participants use a radio band autonomously.
Precise and quick classification enables \ac{CR} participants to efficiently use their radio band without interfering with the transmissions of others.
In addition, an automatic radio signal classification system is also essential for monitoring compliance with the frequency plan.
Due to the large number of different types of radio signals, manual classifications are not economical.
We utilized signal processing and \ac{ML} methods to classify digital operating modes and their parameters in amateur radio scenarios exclusively on spectrograms with computer vision models.
Such models can support spectrum monitoring authorities or organizations identifying band intruders, like the \ac{FCC}.

This work is related to \ac{AMC} studies, interested in classifying the modulations of radio transmissions, e.g., \ac{FSK}, \ac{PSK}, \ac{MFSK} or \ac{OFDM}.
There is another perspective of digital modulation in amateur radio, usually due to hardware limitations, which originated through the requirements of speech transmissions:
The already modulated signal in the \ac{AF} is typically transmitted via the \ac{USB} of a transceiver.
In this case, the \ac{RF} is not generated directly, for example, by a specialized digital transceiver.
This technique allows transmitting a similar modulated signal in the \ac{RF} up to a bandwidth of \(3\) kHz.
In amateur radio, these modulations are seen as operating modes with specific modulation parameters and specified (e.g., error-correction \& synchronization) procedures, such as BPSK31 or Olivia 4/250.
In this example, BPSK (a \ac{PSK} modulation) and Olivia (a \ac{MFSK} modulation) could be differentiated as the pure operating mode and \(31\) and \(4/250\) as parameters for these modes.
This study refers to them as \ac{OM} and \ac{OMP}.
The main contributions of this study are:
\begin{itemize}

    \item Generation of \(98\) different parameterized radio signals from \(17\) digital operating modes \& transmission on the \(70\) cm (UHF) amateur radio band. Recording these transmissions with two different architectures of SDR receivers for real-world evaluation.

    \item Training of the computer vision models ResNet-18, EfficientNetB0, and Vision Mamba Tiny exclusively on spectrograms of limited non-transmitted data, utilizing an online data augmentation pipeline to simulate radio channel impairments.
    
    \item Analysing the impact of varying signal durations and number of FFT bins on classification utilizing real-world radio signals.
    
    \item Evaluation of the models across multiple simulated SNRs and assessment of the simulated channel impairments.
   
\end{itemize}

\section{Related Work}

Current \ac{AMC}/\ac{ML} studies utilized deep learning models such as CNN/LSTM-based models \cite{emam2020comparative} \cite{huynh2020mcnet}, and Transformer-based models \cite{boegner2022largenarrow} \cite{ChannelEst2020}.
Some of them focus on feature engineering \cite{zhao2020automatic} \cite{meng2018automatic}, or data augmentation \cite{huang2019data}, and compare \ac{ML} \& traditional methods \cite{peng2017modulation}.
Multi-tasking approaches have been studied in \cite{chang2021multitask}.
A multi-modal approach for 5G has been presented in \cite{qi2020automatic}.
\ac{ML} approaches not requiring labeled data for training were examined in \cite{davaslioglu2022self} and \cite{chen2021signet}.
Traditional methods can also be used for \ac{AMC}, seen as feature extractors, such as \ac{FFT}-based measurements \cite{scholl2016exact} or wavelet-transform-based baud rate estimations \cite{gao2012baud}.
\ac{ML} methods are also used in digital radio communication, e.g., deep-learning-based channel estimation \cite{ChannelEst2020}.
The challenge of classifying a signal can be extended with simultaneous detection \cite{vagollari2021joint}.
The studies \cite{boegner2022large} \cite{comar2023modulation} also deal with detecting and classifying one or more signals in wideband scenarios.
Most of these studies utilized synthetic data, e.g., the RADIOML 2016.10A dataset \cite{o2016radio}, or generated synthetic data for evaluation \cite{boegner2022largenarrow}.
An exception is \cite{o2018over}, in which the signals were transmitted within a room, and \cite{ya2022large}, which classified real-world ADS-B signals.
This study utilized methods from the fields of telecommunications, \ac{DSP}, \ac{SDR}, and amateur radio.
A comprehensive overview of wireless communication can be found in \cite{Rappaport2002}.
In \cite{Tse2005} and \cite{Rappaport2002}, the impairments that affect radio channels in real environments, such as path loss, fading, and interference, are described in detail.
We used \ac{DSP} algorithms for filtering and processing the recordings of our transmissions, which can be read about in \cite{Lyons2010}.
These are also described more in detail in the context of \ac{SDR} receivers in \cite{Heuberger2017} and \cite{Reed2002}.
An overview of digital modulation techniques can be found in \cite{Proakis2008}.
A detailed explanation of the digital amateur radio modes can be found in \cite{Ford2007}.

\section{Data}

\subsection{Radio Signal Generation \& Overview}

\begin{table}
\centering
\caption{
Overview of our generated signals: \(98\) different operating mode parameters (\ac{OMP}) out of \(17\) operating modes (\ac{OM}).
}
\label{tab_SigOverview}
\begin{tabularx}{0.47\textwidth}{|l|X|}
\hline
\textbf{OM} & \textbf{OMP} \\ \hline
BPSK & 31, 63, 63F, 125, 250, 500, 1000 \\
QPSK & 31, 63, 125, 250, 500 \\
8PSK & 125, 125F, 125FL, 250, 250F, 250FL, 500, 500F, 1000, 1000F, 1200F \\
MC-PSK & 125C12, 250C6, 500C2, 500C4, 800C2, 1000C2 \\
PSKR & 125, 250, 500, 1000 \\
Olivia & 4/125, 4/250, 8/250, 8/500, 16/500, 16/1000, 32/1000, 64/2000 \\
Contestia & 4/125, 4/250, 4/500, 8/250, 8/500, 16/500, 32/1000, 64/2000 \\
MFSK & 4, 8, 11, 16, 22, 31, 64, 64L, 128, 128L \\
DominoEx & EX Micro, EX4, EX5, EX8, X11, X16, X22, X44, X88 \\
Thor & Micro, 100, 11, 16, 22, 25x4, 4, 5, 50x1, 50x2, 8 \\
Throb & BX1, BX2, BX4, OB1, OB2, OB4 \\
MT63 & 500S, 500L, 1000S, 1000L, 2000S, 2000L \\
OFDM & 500F, 750F, 3500 \\
RTTY & RTTY \\
IFKP & IFKP \\
CW & CW \\
Noise & Noise \\ \hline
\end{tabularx}
\end{table}

We generated radio signals of digital modes used in (amateur) radio communication.
These modes include traditional methods, such as CW/Morse; and FSK modes, such as RTTY; more modern modes, like PSK modes, such as BPSK31; and modes with significantly higher bandwidths, such as MFSK and OFDM modes.
Each of these modes is characterized by mode-related parameters that determine, for example, the signal bandwidth, the baud rate, and the size of the modulation alphabet.
As shown in Table \ref{tab_SigOverview}, we generated \(98\) different parameterized radio signals (\ac{OMP}) from \(17\) digital operating modes (\ac{OM}) by utilizing Fldigi\cite{Fldigi2023}, which radio amateurs and emergency services widely use for digital communication.
We generated \(180\)s for training \(D_{Train}\) and \(60\)s for validation \(D_{Val}\) for each \ac{OMP} with random characters as payloads in \ac{AF}.

\subsection{Radio Signal Transmission, Recording \& Postprocessing}
\label{lbl_RTransRecPostProc}

We used distinct Wikipedia article excerpts as payloads for \(D_{Test}\) and generated \(75\)s for each \ac{OMP} for our \ac{USB}-transmission with a Kenwood TS-2000X with about \(5\) Watt.
We recorded these transmissions using an RTL-SDR, a heterodyne receiver, and a HackRF, a direct-conversion receiver.
The RTL-SDR is referred to as R0, and the HackRF as R1.
Both SDR receivers were operated with a sampling rate of \(1\) MHz with a frequency offset of \(200\) kHz related to the transmission frequency, providing complex I/Q baseband representations.
Ground Plane Antennas were used for both the transmitter and the receivers.
R0 and R1 were located in different locations to reduce the risk of local interference, and the transmitter was about \(1\) km away from both.
Due to the urban environment, both receivers had no line of sight with the transmitter.

From the recorded complex I/Q baseband data, a narrowband channel centered at the transmission frequency was extracted for R0 and R1.
Each channel was filtered to a bandwidth of \(3\) kHz (\(6\) kHz sampling rate). The complex I/Q signals were then \ac{USB} demodulated to obtain real-valued representations.
A channel bandwidth of \(3\) kHz was selected to ensure direct compatibility with standard amateur radio equipment.
Therefore, the computation of the spectrograms for the classification can be performed directly on the \ac{AF} output of a transceiver (e.g., via the jack connection) or a WebSDR receiver (via a virtual microphone).
Based on our data generation specifications, we automatically cut and labeled all \(98\) \ac{OMP} per receiver.
These data sets are referred to as \(D_{Test/R0}\) and \(D_{Test/R1}\).
We achieved an \ac{SNR} of approx. \(35\) dB with R0 and \(31\) dB with R1.
After \(4\)h of transmission, a maximum frequency drift of \(14\) Hz was measured at R0 and \(160\) Hz at R1.

\section{Methods}

\subsection{Classification Models}
\label{lbl_ClassModels}

This study focuses on two CNN-based models: \ac{RN-18} (\(11.4\)M params) \cite{he2016deep}, \ac{EN-B0} (\(5.3\)M params) \cite{pmlr-v97-tan19a}, and a Vision Transformer: \ac{Vim-Ti} (\(7\)M params) 
 \cite{zhu2024vision}.
\ac{RN-18} was selected due to its proven effectiveness as a classic CNN model.
\ac{EN-B0} was selected because it can achieve a higher classification accuracy with fewer parameters on ImageNet \cite{deng2009imagenet}.
\ac{Vim-Ti} was included as it represents a state-of-the-art Vision Transformer model.
We selected the tiniest model from each family.
All models were pre-trained on ImageNet \cite{deng2009imagenet}, as preliminary experiments have shown that using non pre-trained models reduced accuracy by an average of \(26\%\).
We considered the spectrograms of our signals as images and used them for all three input channels of our models.
We trained our models with Cross-Entropy loss. 
We applied Early Stopping with a patience of \(5\) to limit the potential risk of overfitting.
As a training algorithm, we applied \ac{SGD} with a learning rate of \(0.001\) and a momentum of \(0.9\).
The batch size was set to 256.
We chose SGD over Adam as the optimizer, as it showed more stable behavior during our training across all models.

\subsection{Online Data Augmentation \& Channel Impairments}

We showed in preliminary experiments with all of our models that training on the limited data set \(D_{Train}\) is not practical if the models are evaluated on real-world transmitted data \(D_{Test/R0}\) and \(D_{Test/R1}\).
One reason is that we only have \(180\)s of data per \ac{OMP} in \ac{AF}, and real-world radio transmissions are affected by various radio channel impairments \cite{Rappaport2002} \cite{Tse2005}. 
We, therefore, utilized an online data augmentation pipeline for the simulation of channel impairments and the expansion of our data for training, including the following augmentations:

\textit{Amplify}: Scales the amplitude of the recording by a specified factor.
\textit{FreqShift}: Shifts the recording frequency by a specified amount in Hz. It applies a complex exponential to the waveform, effectively shifting its frequency components. The shifted signal is filtered to remove the lower sideband, and the result is converted back to a real-valued representation.
\textit{SimTone}: Adds a simulated single-tone interference to the recording. It generates a sine wave at a specified frequency and amplitude. This tone is added to the recording.
\textit{Noise}: Adds Gaussian White Noise to the recording to match a specified \ac{SNR} in dB.
For \(D_{Train}\) and \(D_{Val}\), these augmentations were applied in the following order:
\[
Aug = \{\text{Amplify}, \text{FreqShift}, \text{SimTone}_1, \text{SimTone}_2, \text{Noise}\}
\]
For \(D_{Train}\), this sequence was applied five times, with an additional instance of unaugmented data:%, resulting in a sixfold expansion of the training data:
\[
D_{Train}' = \{ Aug(D_{Train})_1, \dots, Aug(D_{Train})_5, D_{Train} \}
\]
The augmentation parameters were randomly selected each time an augmentation was applied within these ranges: Amplify \( \in [0.1, 2] \), FreqShift \( \in [-500, 500] \), \(\text{SimTone}_{1/2}\) \( \in ([10, 2990]; [0, 0.3]) \), and Noise \( \in [-6, 42] \).
For \(D_{Val}'\) the same augmentation sequence was applied once without unaugmented data and
with fixed parameters: Amplify \(0.5\), FreqShift \(400\), \(\text{SimTone}_1\) \(1000; 0.03\), \(\text{SimTone}_2\) \(2300; 0.015\) and Noise \(30\).

In this way, we provided new unseen training data with random channel impairments at each training epoch.
During training, we validated our models on a fixed set that mimics a concrete real-world scenario.
No augmentation was applied on our real-world test datasets \(D_{Test/R0}\) and \(D_{Test/R1}\).

\section{Experiments \& Results}

\begin{table}
    \centering
    \caption{
    The accuracies with and without data augmentations of our operating mode parameters (OMP) (98 classes) and operating mode (OM) (17 classes) classifications with EfficientNet-B0 (EN-B0), considering a duration of \(2\)s and \(128\) FFT bins.
    Evaluated on our real-world transmissions, recorded by a heterodyne (R0) and a direct-conversion (R1) SDR receiver.
    In the second part of the Table, a specific data augmentation in training was removed to assess its impact on classification.
    }
    \label{tab_Augs}
    \begin{tabular}{|l||c|c||c|c|}
    \hline
        \multirow{2}{*}{\textbf{Augmentation}} & \multicolumn{2}{c||}{\textbf{R0}} & \multicolumn{2}{c|}{\textbf{R1}} \\
        & \textbf{OMP [\%]} & \textbf{OM [\%]} & \textbf{OMP [\%]} & \textbf{OM [\%]} \\ \hline
         without all Augs. & 3.14 & 8.23 & 3.03 & 7.35 \\ 
        with all Augs. & 84.05 & 92.64 & 79.74 & 89.88 \\ \hline
        -Amplify & 78.90 & 87.59 & 58.14 & 64.70 \\ 
        -FreqShift & 83.35 & 92.05 & 42.45 & 57.74 \\ 
        -\(\text{SimTone}_1\) & 83.90 & 92.61 & 81.19 & 90.02 \\ 
        -\(\text{SimTone}_{1/2}\) & 80.67 & 90.57 & 74.41 & 84.75 \\ 
        -Noise & 4.73 & 11.90 & 5.23 & 17.21 \\ \hline
      
    \end{tabular}
\end{table}

Our preliminary experiments showed that training on signals in \ac{AF} is impractical for models evaluated on real-world transmissions influenced by various radio channel impairments; see Table \ref{tab_Augs}.
We conducted a channel impairment study to assess the impact of our simulated impairments on real-world classification.
In this study, we successively excluded all augmentations to evaluate their impact on real-world classification accuracies.
As noted in Table \ref{tab_Augs}, e.g., our \textit{Noise} augmentation proved particularly important for training our models.
\begin{table*}
    \centering
    \caption{
    The accuracies of our operating mode parameters (OMP) (98 classes) and operating mode (OM) (17 classes) classifications.
    Evaluated on our real-world transmissions, recorded by a heterodyne (R0) and a direct-conversion (R1) SDR receiver.
    }
    \label{tab_ResClass}
    \begin{tabular}{|l|l||c|c||c|c||c|c|}
    \hline
        \multirow{2}{*}{\textbf{Dur}} & \multirow{2}{*}{\textbf{n-FFT}} & \multicolumn{2}{c||}{\textbf{EfficientNetB0 (EN-B0)}} & \multicolumn{2}{c||}{\textbf{ResNet-18 (RN-18)}} & \multicolumn{2}{c|}{\textbf{Vision Mamba Tiny (Vim-Ti)}} \\ 
        & & \textbf{OMP [\%]} & \textbf{OM [\%]} & \textbf{OMP [\%]} & \textbf{OM [\%]} & \textbf{OMP [\%]} & \textbf{OM [\%]} \\ \hline
        4s & 256 & \textbf{85.47 ± 0.94} & \textbf{93.80 ± 0.75} & 81.58 ± 3.69 & 90.48 ± 2.85 & \textbf{84.40 ± 2.27} & \textbf{92.91 ± 2.01} \\ 
        4s & 128 & 84.89 ± 0.87 & 93.46 ± 0.52 & \textbf{84.21 ± 2.38 } & 92.51 ± 1.79 & 80.83 ± 4.31 & 89.94 ± 3.48 \\ 
        4s & 64 & 81.02 ± 4.33 & 89.51 ± 4.11 & 82.99 ± 4.67 & \textbf{93.12 ± 1.80} & 81.08 ± 4.18 & 90.20 ± 3.41 \\ \hline
        3s & 256 & \textbf{83.81 ± 1.31} & 92.20 ± 1.43 & 79.24 ± 4.13 & 88.91 ± 3.51 & \textbf{82.93 ± 2.04} & 92.40 ± 1.30 \\ 
        3s & 128 & 83.73 ± 1.34 & \textbf{92.60 ± 0.98} & 81.81 ± 2.15 & 90.25 ± 2.19 & 80.92 ± 4.03 & 89.34 ± 3.76 \\ 
        3s & 64 & 82.35 ± 3.02 & 90.46 ± 3.20 & \textbf{82.10 ± 3.86} & \textbf{91.80 ± 2.55} & 82.64 ± 2.95 & \textbf{92.42 ± 1.43} \\ \hline
        2s & 256 & 81.67 ± 2.01 & 90.39 ± 1.98 & 77.40 ± 4.65 & 87.02 ± 4.42 & 81.22 ± 1.86 & 90.72 ± 1.72 \\ 
        2s & 128 & \textbf{81.90 ± 2.16} & \textbf{91.26 ± 1.38} & \textbf{81.06 ± 2.27} & 89.92 ± 2.01 & 78.28 ± 4.53 & 88.57 ± 3.67 \\ 
        2s & 64 & 81.17 ± 2.86 & 90.51 ± 2.39 & 79.78 ± 5.15 & \textbf{90.21 ± 2.57} & \textbf{81.94 ± 1.89} & \textbf{90.72 ± 1.72} \\ \hline
        1s & 256 & 77.48 ± 2.44 & 87.23 ± 2.34 & 73.12 ± 3.73 & 83.82 ± 3.75 & \textbf{76.89 ± 3.06} & 86.89 ± 2.59 \\ 
        1s & 128 & 77.62 ± 3.05 & \textbf{87.81 ± 2.32} & \textbf{76.30 ± 1.55} & \textbf{86.61 ± 1.84} & 76.62 ± 4.17 & 86.70 ± 3.71 \\ 
        1s & 64 & \textbf{77.63 ± 3.94} & 87.16 ± 3.14 & 74.33 ± 4.67 & 85.92 ± 2.64 & 77.75 ± 2.45 & \textbf{87.93 ± 1.94} \\ \hline
    \end{tabular}
\end{table*}
This study also investigated the impact of varying signal durations and the number of FFT bins on classification evaluated on our real-world datasets.
To explore this, we conducted experiments with our three models with separate training and evaluation for each combination across the signal durations of \(1\)s, \(2\)s, \(3\)s, \(4\)s, and \(64\), \(128\), \(256\) FFT bins for spectrogram computation.
The results of these experiments are detailed in Table \ref{tab_ResClass}.
\begin{figure}
	\centerline{\includegraphics[width=0.6\linewidth]{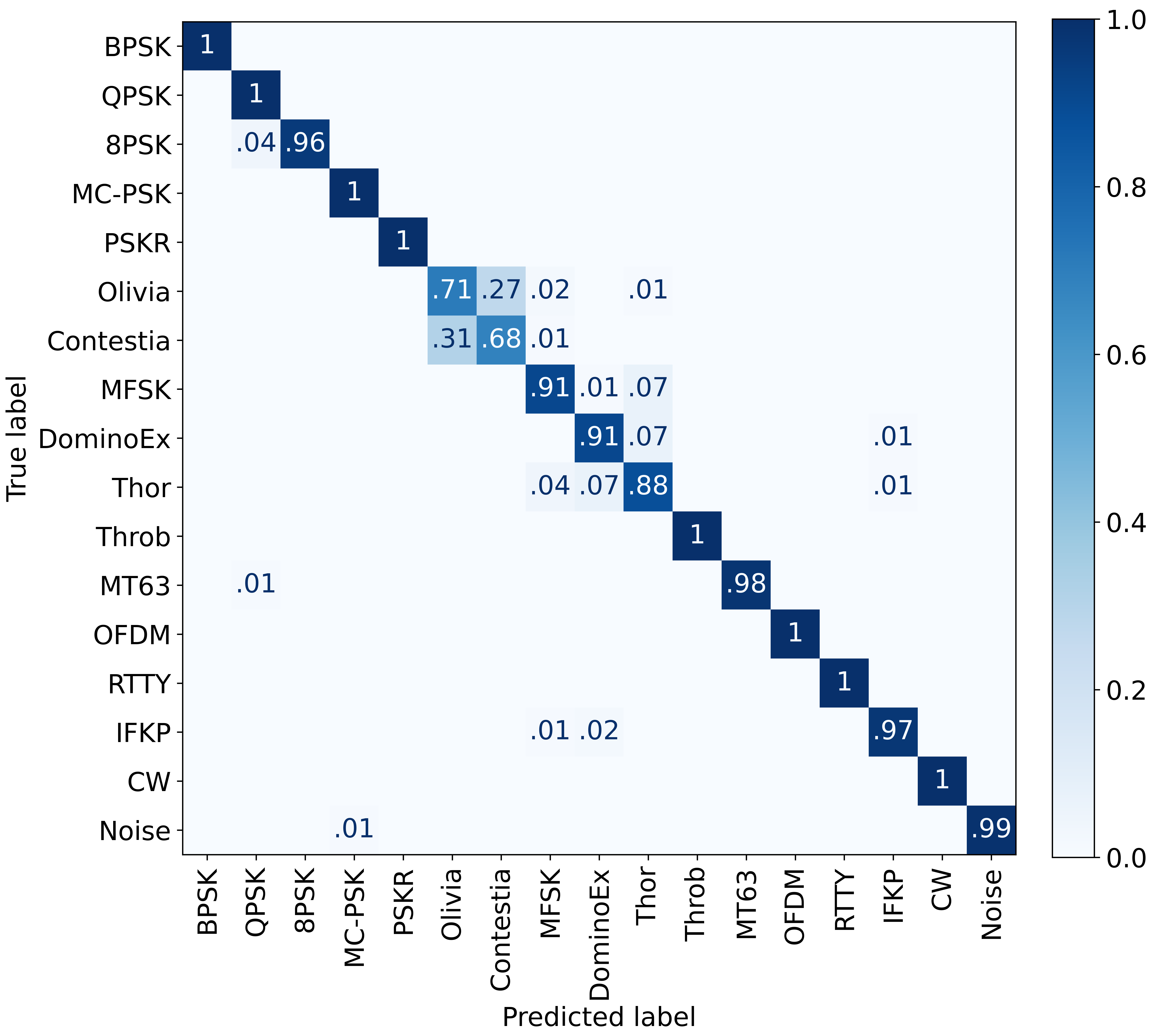}}
         \caption{
            This normalized confusion matrix visualizes the operating mode (\ac{OM}) confusions of our best model, EfficientNet-B0 (EN-B0), considering a duration of \(2\)s and \(128\) FFT bins.
            Evaluated for both receivers R0 and R1.
            }
	\label{fig_Confusion}
\end{figure}

\ac{EN-B0} performed the best of our models with an accuracy of \(93.80\%\) across the \(17\) \ac{OM} and \(85.47\%\) across all \(98\) \ac{OMP}.
The accuracy significantly increased when the signal was analyzed for at least \(2\)s for classification.
We aimed to minimize the signal duration while maintaining meaningful classification, enabling higher robustness through multiple decisions in real-world scenarios.
Therefore, we selected our best model, \ac{EN-B0}, for further evaluation in this study, considering a duration of \(2\)s.
Figure \ref{fig_Confusion} shows that most confusion occurred between Contestia and Olivia.
\begin{figure}
	\centerline{\includegraphics[width=0.80\linewidth]{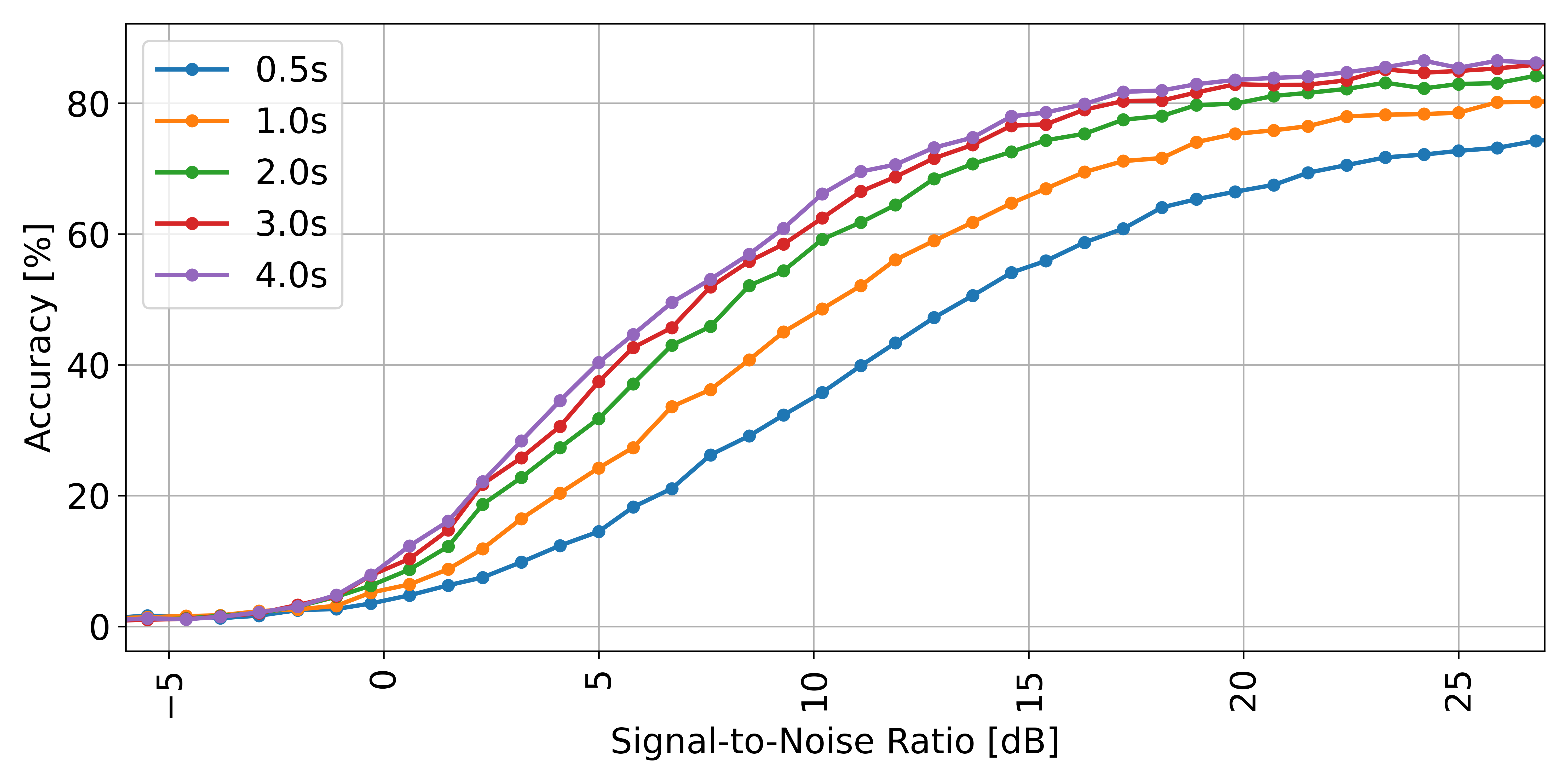}}
	\caption{
        The operating mode parameters (\ac{OMP}) classification accuracies of our best model, EfficientNet-B0 (EN-B0), utilizing \(128\) FFT bins across multiple simulated SNRs.
        }
	\label{fig_SNR}
\end{figure}
Finally, we evaluated our best model \ac{EN-B0} across multiple simulated \acp{SNR}.
Figure \ref{fig_SNR} presents the classification accuracy of \ac{EN-B0} with \(128\) FFT bins across \(-6\) and \(27\) dB with varying durations.

We trained our models intending to classify \ac{OMP} and evaluate our models with the datasets \(D_{Test/R0}\) and \(D_{Test/R1}\) from the receivers R0 and R1 with both the classification task \ac{OMP} and \ac{OM}.
To ensure comparability, we trained separate models for each possible combination of varying durations and the number of FFT bins.
To ensure comparability across models trained with varying durations, we extracted the same amount of classification windows from \(D_{Test/R0}\) and \(D_{Test/R1}\) using a consistent shift of \(0.5\)s, maintaining an equal number of evaluation decisions for all configurations.
In our case, \(75\)s per \ac{OMP}, which means up to \(150\) decisions are made per \ac{OMP} and receiver for evaluation. With \(98\) different \ac{OMP}, this means \(14.700\) decisions per receiver.

\section{Discussion}

It can be noted that \textit{Noise} augmentation was the most important of all our augmentations; see Table \ref{tab_Augs}.
The \textit{FreqShift} augmentation was particularly important for R1, as this receiver was also affected by a higher frequency drift; see Section \ref{lbl_RTransRecPostProc}.
The receivers were also influenced by unavoidable radio interference, which is why the (multiple) application of \textit{SimTone} can be helpful for this scenario.
This local interference also constantly influences the gain control of SDR receivers, and the real-world signals are typically received with different \ac{SNR}, which is where \textit{Amplify} can support.
As shown in Table \ref{tab_ResClass}, \ac{EN-B0} performed the best of our models.
It can be noted that the duration of a signal was more important than the choice of the number of FFT bins for classification.
The accuracy significantly increased when the signal was analyzed for at least \(2\)s for classification.
Most of the confusion occurred within an \ac{OM} and between Contestia and Olivia; see Figure \ref{fig_Confusion}.
Contestia is a modification of Olivia that has been specially adapted for higher speeds and more robust transmissions under the conditions of amateur radio competitions.
For example, the encoding has been optimized for transmission speed by only allowing the transmission of capital letters.
The transmitted signal, therefore, does not differ in most parts. 
The \acp{OM} MFSK, DominoEx, and Thor are similarly close to each other. 
Figure \ref{fig_SNR} shows that a longer signal resulted in a more meaningful classification even at a poorer \ac{SNR}.
Above approx. 25 dB \ac{SNR}, the accuracy no longer changed significantly for the better.
The methods employed and the design decisions, such as selecting a channel width of \(3\) kHz, ensure that our approach is lightweight \& compatible with standard amateur radio equipment and WebSDR receivers for direct integration; see Sections \ref{lbl_RTransRecPostProc} and \ref{lbl_ClassModels}.

\section{Conclusion}

With limited non-transmitted data (\(180\)s) per \ac{OMP} in \ac{AF} with random characters as payloads, it was possible to train our three ML models exclusively on spectrograms successfully.
By evaluating our models on real-world transmissions, we showed that adding artificial noise across multiple SNR conditions and applying a frequency shift augmentation was necessary for successful training.
Our best model, EfficientNetB0, achieved an accuracy of \(93.80\%\) across the \(17\) \ac{OM} and \(85.47\%\) across all \(98\) \ac{OMP}.
In general, it can also be noted that longer signal durations significantly enhanced classification accuracy, especially under challenging \ac{SNR} conditions.
In future studies, we will analyze the differences in classification at real-world poor \ac{SNR} compared to the simulated ones.

\section*{Acknowledgment}

We would like to thank the TH Nürnberg for using their amateur radio station DF0OHM and Prof. Dr. Thomas Lauterbach for his advice on radio topics.
This study was supported by the Bavarian Collaborative Research Program (Bayerischen Verbundforschungsprogramms, BayVFP) under grant ID DIK0517/02.

\newpage

% \section*{References}

\bibliographystyle{IEEEtran}
\bibliography{IEEEabrv,conference_101719}

\end{document}